\pdfoutput=1
\documentclass[11pt]{article}

\usepackage[]{acl}

\usepackage{times}
\usepackage{graphicx}
\usepackage{latexsym}
\usepackage{booktabs}
\usepackage[normalem]{ulem}
\usepackage{threeparttable}
\usepackage{tablefootnote}
\usepackage{subfigure}
\usepackage{caption}
\usepackage{multirow}
\usepackage[T1]{fontenc}
\usepackage[utf8]{inputenc}

\usepackage{microtype}
\usepackage{amsmath, amsfonts, amssymb}
\title{On Synthetic Data
for Back Translation\thanks{\hspace{2pt} This work was done during the internship of the first author at Tencent AI Lab. The code is available at \url{https://github.com/Jiahao004/Data-for-BT}}}

\author{Jiahao Xu$^1$, Yubin Ruan$^2$, Wei Bi$^2$, Guoping Huang$^2$, \\\bf{Shuming Shi$^2$, Lihui Chen$^1$, Lemao Liu$^2$} \\
$^1$Nanyang Technological University, $^2$Tencent AI Lab \\
$^1$\texttt{\{jiahao004@e.,elhchen@\}ntu.edu.sg} \\
$^2$\texttt{\{brantruan,victoriabi,donkeyhuang,shumingshi,redmondliu\}} \\\texttt{@tencent.com}
}

\begin{document}
\maketitle
\begin{abstract}
% Previous work has proven that back-translation is an effective method for advancing the performance of modern neural machine translation (NMT) models. Despite various contributions made for heuristic back-translation strategy in both synthetic corpus generation and forward model training, the inner mechanism of back-translation as well as mathematical explanation has not been explored yet. Our motivation in this work is to mathematically explain the back-translated synthetic data on NMT training objective, and find optimized strategy for synthetic potential candidates selection. 

Back translation (BT) is one of the most significant technologies in NMT research fields. Existing attempts on BT share a common characteristic: they employ either beam search or random sampling to generate synthetic data with a backward model but seldom work studies the role of synthetic data in the performance of BT. This motivates us to ask a fundamental question: {\em what kind of synthetic data contributes to BT performance?}
Through both theoretical and empirical studies, we identify two key factors on synthetic data controlling the back-translation NMT performance, which are quality and importance. Furthermore, based on our findings, we propose a simple yet effective method to generate synthetic data to better trade off both factors so as to yield a better performance for BT. We run extensive experiments on WMT14 DE-EN, EN-DE, and RU-EN benchmark tasks. By employing our proposed method to generate synthetic data, our BT model significantly outperforms the standard BT baselines (i.e., beam and sampling based methods for data generation), which proves the effectiveness of our proposed methods.
\end{abstract}

\section{Introduction}
Since the birth of neural machine translation (NMT)~\cite{bahdanau2014neural,sutskever2014sequence} back translation (BT)~\cite{sennrich-etal-2016-improving} has quickly become one of the most significant technologies in natural language processing (NLP) research field. This is because 1) it provides a simple yet effective approach to advance the supervised NMT by leveraging monolingual data~\cite{edunov2018understanding} and it also serves as a key learning objective in unsupervised NMT~\cite{artetxe2018unsupervised,lample2018phrase}; 2)  back translation even plays a significant role in other NLP research fields beyond translation such as paraphrasing~\cite{mallinson2017paraphrasing} and style transfer~\cite{prabhumoye2018style,zhang2018style}.
%\bi{as a general data augmentation method?}

Back translation consists of two steps, namely synthetic corpus generation with a backward model and parameter optimization for the forward model. Various contributions have been made on improving back translation, for instance, iterative back translation~\cite{hoang2018iterative}, tagged back translation~\cite{caswell2019tagged}, confidence weighting~\cite{wang2019improving}, data diversification~\cite{nguyen2020data}. Although these efforts differ in some aspects, all of them share a common characteristic: they employ a default way to generate synthetic data in the first step of BT which is either beam search or random sampling with a backward model. 
Seldom work studies the consequences of synthetic corpus to back translation and hence it is unclear how synthetic data influences the final performance of BT. 

The early study empirically suggests the quality of the synthetic corpus is vital for BT performance~\cite{sennrich-etal-2016-improving}. However, recent studies illustrate better test performance can be achieved by low quality synthetic corpus~\cite{edunov2018understanding}. This contradictory observation indicates the quality of synthetic data is not the only element that affects the BT performance. Hence, this fact naturally raises a fundamental question: {\em what kind of synthetic data contributes to back translation performance?}

In this paper, we attempt to take a step forward toward the above fundamental question. To this end, we start from a critical objective in semi-supervised learning, which is defined by the marginal distribution of a target language. Then we derive an approximate lower bound of the objective function, which is closely related to the objective of back translation. Corresponding to this lower bound, we theoretically find two related elements for maximizing such a lower bound: quality of synthetic bilingual data and importance weight of its source. Since both elements are mutually exclusive to some extent,  it may induce contradictory observation if one judges the BT performance according to a single element. In addition, such a theoretical explanation is supported by our empirical experiments. Furthermore, based on our findings, we propose a new heuristic approach to generate synthetic data whose both elements are better balanced so as to yield improvements over both sampling and beam search based methods. Extensive experiments on three WMT14 tasks show that our BT consistently outperforms the standard sampling and beam search based baselines by a significant margin. 

Our contributions are three folds:
\begin{enumerate}
    \item We point out that importance weight and quality of synthetic candidates are two key factors that affect the NMT performance. 
    \item We propose a simple yet effective method for synthetic corpus generation, which could better balance the quality and importance of synthetic data.
    \item Our experiments prove the effectiveness of the aforementioned strategy, it outperforms beam or sampling decoding methods on three benchmark tasks.
    
\end{enumerate}

\section{Revisiting Back Translation}
NMT builds a probabilistic model $p(y|x; \theta)$ with neural networks parameterized by $ \theta$, which is used to translate a sentence $x$ in source language $\mathcal{X}$ to a sentence $y$ in target language $\mathcal{Y}$. 
%In NMT, one is always interested in translating a source sequence $x\in \mathcal{X}$ to a target sequence $ y \in \mathcal{y}$.  
The standard wisdom to train the model is to minimize the following objective function over a given bilingual corpus $\mathcal{B}=\{( x_i,  y_i)\}$:
\begin{equation}
\label{nmt_obj}
    \ell(\mathcal{B};\theta)=\sum_{ (x_i, y_i)\in \mathcal{B}}\log{p(y_i|x_i; \theta)}
\end{equation}

Recently ~\citet{sennrich-etal-2016-improving} propose a remarkable method called {\bf Back Translation (BT)} to improve NMT by using a monolingual corpus $\mathcal{M}$ in target language $\mathcal{Y}$ besides $\mathcal{B}$ and 
back translation becomes one of the most successful techniques in NMT~\cite{fadaee2018back,edunov2018understanding}. %The key idea of back translation is to generate a synthetic corpus for $\mathcal{M}$ and then optimize $$ 
At a high level, back translation can be considered as a semi-supervised method because it leverages both labeled and unlabeled data. 
Suppose $p(x|y; \pi)$ is the backward translation model whose parameter $\pi$ is optimized over $\mathcal{B}$, the key idea of back translation can be summarized as the following two steps:
\begin{itemize}
    \item {\bf Synthetic Corpus Generation:} It firstly back-translates each target sentence $y\in \mathcal{M}$ to $\hat{x}$ obtain a synthetic bilingual corpus $\{(\hat{x}, y)\mid y \in \mathcal{M}\}$ by $p(x|y; \pi)$.
    \item {\bf Parameter Optimization:} It combines both authentic corpus $\mathcal{B}$ and the synthetic corpus and then optimizes the parameter $\theta$ by minimizing the loss 
    \begin{equation}
    \ell(\mathcal{B}; \theta) + \sum_{y \in \mathcal{M}} \log p(y|\hat{x}; \theta) 
    \label{eq:bt-loss}
    \end{equation}
\end{itemize}
\noindent To make BT more efficient, the standard configuration is widely adopted: each sentence $y$ is required to generate a single source $\hat{x}$ and both two steps are performed for a single pass. We follow this standard in this paper for generality but our idea in this paper is straightforward to apply to other configurations such as~\cite{gracca2019generalizing,hoang2018iterative,nguyen2020data}. 

In the first step, there are two main strategies to generate the synthetic corpus, i.e., deterministically decoding and randomly sampling with $p(x|y; \pi)$.
% This training objective can be established on bilingual corpus as long as this corpus is parallel and authentic in each language space. However, such authentic bitext is naturally limited and one of the solutions lies on back-translating monolingual target languages into source side, thus to generate synthetic language pairs as a supplement. Large contribution has been made for producing high-quality synthetic corpus, in other words, the candidates' posterior $\log p(y|x)$, approximately estimated by a back translation model $\log p(x|y;\pi)\sim \log p(y|x)$, is relatively high. 
The first strategy aims to search the best candidate as follows,
\begin{equation}
    \hat x^{b}=\arg \max p(\hat x|y; \pi)
\end{equation}
\noindent The above optimization is achieved by the beam search decoding, which can be regarded as a degenerated shortest path problem with respect to the $\log p(\hat x|y; \pi)$ with limited routing attempts.
The alternative strategy is random sampling: it randomly samples a token with respect to the distribution estimated by a back-translation model at each decoding step. Such a process can be modelled by,
\begin{equation}
    \hat{x}^{s}=\texttt{rand}\{p(\hat x| y;\pi)\}
\end{equation}

\paragraph{Research Question}
Prior work points out~\cite{sennrich-etal-2016-improving} that the synthetic corpus with high quality is beneficial to the final performance of back translation. 
However, the recent studies \citep{edunov2018understanding} find that NMT models with unsatisfactory BLEU score corpus, for instance, the corpus generated by sampling based strategy, also establish the state-of-the-art (SOTA) achievement among back-translation NMT models. 

This contradictory fact indicates that the quality of synthetic corpus is not the sole element for back translation. 
This motivates us to study a fundamental question for back translation: {\em what kind of synthetic corpus is beneficial to back translation?}

\section{Understanding Synthetic Data by Two Factors}

To answer the fundamental question presented in the previous section, we first start from the marginal likelihood objective defined on the target language $\mathcal{Y}$, and then we theoretically explain two factors (i.e., quality and importance) that are highly related to the training objective of back translation. Finally, we empirically explain why synthetic corpus with low quality may lead to better performance than synthetic corpus with high quality by measuring both factors. 

\subsection{Theoretical Explanation}
Maximizing marginal likelihood is an important principle to leverage unlabeled data. Therefore, 
we rethink back translation from this principle because it makes use of target monolingual corpus $\mathcal{M}$.
For each $y\in \mathcal{M}$, the marginal likelihood objective can be derived by the Bayesian Equation (\ref{eq:total_prob}), Jansen Inequality (\ref{eq:jansen}), and importance sampling (\ref{equ:lb}) as follows:
%\bi{can number the corresponding line and ref in the sentence?}
\begin{align}
   \log p(y; \theta)&=  \log\sum_x p(x)p(y|x; \theta)
   \label{eq:total_prob} \\
    &\geq  \sum_x p(x)\log p(y|x; \theta)
    \label{eq:jansen}\\
    &= \sum_x p(x|y) \frac{p(x)}{p(x|y)}\log p(y|x; \theta) \nonumber\\
    &= \mathbb{E}_{\hat{x}\sim p(\cdot|y)}\Big\{\frac{p(\hat{x})}{p(\hat{x}|y)}\log p(y|\hat{x}; \theta)\Big\} \nonumber\\
    & \approx \frac{p(\hat{x})}{p(\hat{x}|y)} \log p(y|\hat{x}; \theta) 
    \label{equ:lb}
\end{align}
\noindent where $p(x)$ is a language model on source language $\mathcal{X}$, $p(x|y)$ is a backward translation model from $\mathcal{Y}$ to $\mathcal{X}$ which serves as the proposal distribution for importance sampling, and $\hat{x}$ is sampled from $p(x|y)$. If $p(x|y)$ is set as the backward model $p(x|y; \pi)$ optimized on $\mathcal{B}$, the last term in Equation \ref{equ:lb} is the same as the second term in BT loss (i.e., $\log p(y|\hat{x})$ in Eq.~ \ref{eq:bt-loss}),  and the unique difference is the multiplicative term called {\em importance} weight:
%\bi{can add a sentence to describe this the denominator and numerator intuitively }
\begin{equation}
\label{scaling_term}
    \mathrm{Imp}(\hat{x}; y)=\frac{p(\hat{x})}{p(\hat{x}|y)}
\end{equation}
The denominator is the candidate conditional probability to target, and the numerator is the candidate distribution on source language distribution.
Since $\mathrm{Imp}(\hat{x}; y)$ is constant with respect to the parameter $\theta$, maximizing $\log p(y|\hat{x}; \theta)$ in BT loss implicitly maximizes $\mathrm{Imp}(\hat{x}; y) \log p(y|\hat{x})$, which indicates that back translation aims to implicitly maximize the marginal likelihood objective. 
% \bi{this analysis is for per sample loss, right? a bit strange since it seems do not hold for batch loss.}
More importantly, according to Equation \ref{equ:lb} we can find that the following two factors are critical to influence the marginal likelihood $\log p(y; \theta)$:
\begin{itemize}
    \item {\bf Factor 1:} The quality of $\hat{x}$ as a translation of $y$ corresponding to the $\log p(y|\hat{x}; \theta)$ in Eq.~\ref{equ:lb}. 
    \item {\bf Factor 2:} The importance of $\hat{x}$ as a translation of $y$ corresponding to $\mathrm{Imp}(\hat{x}; y)$ in Eq.~\ref{equ:lb}. 
\end{itemize}
Theoretically, if $\hat{x}$ is of higher quality and contains more semantic information in $y$, $p(y|\hat{x}; \theta)$ would be higher and thus it would lead to a higher $\log p(y; \theta)$, which is well acknowledged by prior work~\cite{sennrich-etal-2016-improving,wang2019improving}. In particular, if $\hat{x}$ is with higher importance weight, maximizing $\log p(y|\hat{x}; \theta)$ is more helpful to maximize $\log p(y; \theta)$. On the contrary, if 
$\mathrm{Imp}(\hat{x}; y)$ is very small, it needs to avoid such a sample $\hat{x}$ from $p(x|y)$, which is essentially the rejection control strategy in importance sampling theory~\cite{liu1998rejection,liu2001monte}. 

Unfortunately, in practice, both factors are mutually exclusive to some extent: if $\hat{x}$ is with high quality, $p(\hat{x}|y; \theta)$ would be higher 
as well leading to lower importance weight. This fact can explain the contradictory observation in Sec~2 that BT with high-quality synthetic data sometimes leads to better testing performance, while it may deliver worse performance at other times, which will be later justified in Sec~3.2.

\paragraph{Estimating Two Factors}
To measure the quality of $\hat{x}$ for each $y$, it is natural to use the evaluation metric such as BLEU if the reference translation $x$ of $y$ is available. Otherwise, as a surrogate, we use the log likelihood $\log p(\hat{x}|y; \pi)$ of the backward translation model $\pi$ which is trained on the authentic data $\mathcal{B}$. Similarly, in order to estimate the importance of $\hat{x}$, we train an additional language model $p(x; \omega)$ with GPT~\cite{radford2018improving} on a large monolingual corpus for $\mathcal{X}$. In this way, the importance weight is estimated by
\begin{align*}
    \mathrm{Imp}(\hat x)\approx  \frac{p(\hat x;\omega)}{p(\hat x|y; \pi)}\nonumber
\end{align*}

\subsection{Empirical Justification\label{sec:justification}}
In this subsection, we aim to justify the following statements: 1) encouraging the quality of synthetic corpus may to some extent hurt the performance of BT due to the decrease of importance; 2) judging the testing performance in terms of quality only may be dangerous while it would be meaningful to judge the testing performance by taking into account both factors rather than either factor. To this end, we run some quick experiments on WMT14 datasets whose settings will be shown in Sec~5 later. 

\begin{table}
\centering
\resizebox{\linewidth}{!}{%
\begin{tabular}{lcccc} \toprule
\multicolumn{1}{c}{\textbf{Systems}} & \multicolumn{1}{c}{\textbf{BLEU}($\hat{x}$)} & \multicolumn{1}{c}{$\log p(\hat{x}|y,\pi)$} & \multicolumn{1}{c}{\textbf{Imp}.} & \textbf{Test BLEU}  \\ \midrule
beam                    & \textbf{27.20}                       & -\textbf{15.65}                 & -95.13                            & 32.7                 \\
sampling                 & 7.70                                 & -157.62                         & -\textbf{41.86}                   & 34.1                \\
beam*                   & \textbf{18.50}                       & -\textbf{26.66}                 & -95.07                            & 31.6               \\
% sampling*                & 5.20                                 & -258.85                         & \textbf{32.56}                    & 30.9                \\
\hline
\end{tabular}
}
\begin{tablenotes}
\small
    \item * The checkpoint of the backward model for generating synthetic corpus are only trained for 1 epoch. However, its $\log p(\hat{x}|y,\pi)$ is still measured by a standard backward model $\pi$. 
\end{tablenotes}
\caption{\label{tab:existence} Testing BLEU (on test set), quality (measured by both BLEU and $\log p(\hat x|y;\pi)$) and importance (Imp.) estimation of synthetic data (on dev set) with beam search or random sampling on WMT14 DE-EN task.}
\end{table}

\begin{table}
\centering
\resizebox{\linewidth}{!}{%
\begin{tabular}{lcccc} \toprule
\multicolumn{1}{c}{\textbf{Systems}} & \multicolumn{1}{c}{\textbf{BLEU}($\hat{x}$)} & \multicolumn{1}{c}{$\log p(\hat{x}|y,\pi)$} & \multicolumn{1}{c}{\textbf{Imp}.} & \textbf{Test BLEU}   \\ \midrule
en-de(en)\_beam                    & \textbf{31.90}                       & -\textbf{15.29}                 & -91.07                            & 29.7                \\
en-de(en)\_sampling                 & 10.90                                & -139.71                         & -\textbf{46.88}                   & 30.0                \\
ru-en(ru)\_beam                    & \textbf{33.10}                       & -\textbf{15.49}                 & -89.71                            & 35.9                \\
ru-en(ru)\_sampling                 & 9.50                                 & -155.82                         & -\textbf{47.47}                   & 35.6                \\ \bottomrule
\end{tabular}
}
\caption{\label{tab:existence1}Testing BLEU (on test set), quality (measured by both BLEU and $\log p(\hat x|y;\pi)$) and importance (Imp.) estimation of synthetic data (on development set) with beam search or random sampling on WMT14 EN-DE and RU-EN tasks.}
% \label{tab:existence}
\end{table}

We set up two back translation systems with two different options (i.e., beam search and sampling) to generate synthetic corpus by using the best checkpoint of $p(\hat x|y; \pi)$ tuned on the development set. Both beam search and sampling based BT systems are denoted by beam and sampling. 
In addition, we pick another checkpoint of $p(\hat x|y; \pi)$ which is trained for only 1 epoch, and we use this weak checkpoint to set up another beam search based BT system, which is denoted as beam*. 
Table~\ref{tab:existence} shows BLEU on test dataset, the quality and importance on the development set according to three systems on WMT14 DE-EN task. 

In Table~\ref{tab:existence}, beam is better than sampling in the quality of synthetic corpus but its testing performance is worse. This is meaningful because the former relies on the synthetic corpus with lower importance weight according to our theoretical explanation. In addition, when comparing beam with beam*, we can find that beam delivers better testing performance because its quality is better meanwhile its importance weight is almost similar to that of beam*.  Table~\ref{tab:existence1} consistently demonstrates that it is meaningless to take into account quality only when evaluating BT. These facts justify our statements and provide an answer to the fundamental question in section 2. 

\section{Improving Synthetic Data for BT}
As shown in the previous section, both importance and quality of synthetic corpus are beneficial to the overall testing performance of back translation. It is a natural idea to promote both factors when generating synthetic corpus such that running BT on such corpus leads to better testing performance. However, this is difficult because both factors are mutually exclusive as discussed in Section 3. In this section, we instead propose two methods (namely data manipulation and gamma score) to trade off both factors in the hope to yield better BT performance.

\subsection{Data Manipulation}
Since the synthetic data in sampling based BT is of high importance yet low quality whereas the case for the synthetic data in beam search based BT is opposite,  
we propose a data manipulation method to trade off importance and quality by combining both synthetic datasets. Through balancing the ratio between beam and sampling based synthetic corpora, we expect to find an optimized beam/sampling ratio to further improve NMT model performance. 

Specifically, we randomly shuffle $\mathcal{M}$ and divide it into two parts with the first part accounting for $\gamma$ ($0< \gamma < 1$); then we generate translations for the first part with beam search while generating translations for the second part with sampling. Formally, we use the following corpus $\mathcal{M}^{c}$ as the synthetic corpus for BT: 
\begin{align*}
    \mathcal{M}^{c}=\{(\hat{x}_i^{b},y_i)&_{i=0}^k\}\cup\{(\hat{x}_j^{s},y_j)_{j=k}^{|\mathcal M|}\}\\
    k=&\lfloor \gamma|\mathcal M| \rfloor
\end{align*}
\noindent Where $\hat{x}^b$ denotes a translation of $y$ generated by $p(x|y;\pi)$ with beam search and $\hat{x}^s$ is a translation with sampling, $|\cdot|$ means the size of the corpus, and $\gamma$ is the combination ratio for beam and sampling synthetic corpora. By tuning $\gamma$ here, one can modify the weightage for the number of beam and sampling sentences, to improve back-translation performance by training models on a combined synthetic corpus.

Although this method is easy to implement, its limitation is obvious. Since each $\hat{x}$ is either from beam search or from sampling, the quality of $\mathcal{M}^{c}$ is generally worse than that of beam search and its importance weight is generally worse than that of sampling. 
Consequently, we propose an alternative method in the next part of this section. 
%The back-translation NMT model performance is improved by trading-off the ratio of beam and sampling. However, shortages of this proposed strategy are: hyperparameter search for combining ratio is required since it is sensitive for experiment settings, corpus scale, etc., ; \lemao{The limitations: it can not obtain high importance weight: the importance is worse than sampling and quality worse than beam search}

\subsection{Gamma Score}
% Since we have identified that beam decoded candidates are of high quality, sampling decoded candidates are of high importance, and the Pareto Optimality exists between importance and sampling, 
% In this subsection, we raise such a question, could we directly sample candidates with balanced quality and importance to satisfy the back-translation NMT models requirement, instead of combining beam and sampling corpus?

% It should be noticed that, exact decoding methods are beyond the range of this research, and we only consider the possibility to trade-off the importance and quality of candidates when we sample pseudo sources, thus to increase the back-translation NMT model performance. Therefore, to search as many the pseudo sources in source language space as possible, we employ the sampling decoding method for candidates generation.

The key idea to the alternative method is that it employs a score that balances both quality and importance to generate a translation $\hat x$ for each $y\in \mathcal{M}$. A natural choice of such a score is defined by the interpolation score as follows:
\begin{align*}
    \gamma \log \mathrm{Imp}(\hat x;\omega,\pi)+(1-\gamma)\log p(\hat x|y;\pi)
\end{align*}
\noindent where $\gamma$ is used to trade off both factors as in corpus manipulation. With the help of this score, one may optimize the $\hat x$ by beam search whose interpolation score is the best among all possible translations of $y\in \mathcal{M}$. Unfortunately, such an implementation leads to limited performance in our preliminary experiments, due to two major challenges.

On one hand, the estimations of quality and importance weight of $\hat x$ are not well calibrated, and in particular, quality and importance are mutually exclusive as mentioned before. As a result, beam search with the interpolation score over the exponential space can not guarantee a desirable translation $\hat{x}$ for each $y$.  
On the other hand, quality and importance weight of $\hat x$ are not at the same scale for different $y$, it is difficult to balance both factors with a fixed $\gamma$ in the interpolation score for different $y$. 

To alleviate these issues, we propose a simple method as follows.  
Specifically, firstly, instead of beam search with the interpolation score, we simply utilize the backward translation $p(x|y; \pi)$ to randomly sample a set of candidate translations which is denoted by $A(y)=\{\hat{x}_i\}_i^N$ ($N=50$ in this paper as it works well). ~\footnote{ $N$-best decoding strategy with $p(x|y; \pi)$ to generate $N$ candidates may be another solution which remains as future work.} Then we pick a $\hat x_j$ among $A(y)$ according to the balancing score. Secondly, for each $\hat x$, we normalize the log values of importance and quality of each candidate by its sequence length, then normalize these values with respect to all $N$ candidates as follows:
\begin{align}
    \mathcal{\tilde F}(\hat x_i)=\frac{\log \big(\mathcal{F}(\hat x_i)\big)/\texttt{len}(\hat x_i)- \mu_{\mathcal{F}}}{ \sigma_{\mathcal{F}}}
    \label{eq:norm}
\end{align}
where $\mathcal{F}$ is either importance weight or quality estimations, and $\mu_{\mathcal{F}}=\frac{1}{N}\sum_i \log \mathcal{F}(\hat x_i)$ and $ \sigma_{\mathcal{F}}= \frac{\sum_i(\log \mathcal{F}(\hat x_i)-\mu_{\mathcal{F}})^2}{N-1}$ are mean and variance of $N$ sampled candidates with length normalized. Finally, the Gamma score is defined on the normalized values of importance and quality as follows:
\begin{multline}
         \Gamma(\hat x_i; \omega,\pi) = \\
    \frac{\exp\big(\gamma \tilde{\mathrm{Imp}}(\hat x_i;\omega,\pi)+(1-\gamma) \tilde{p} (\hat x_i|y,\pi) \big)} {\sum_j \exp \big( \gamma \tilde{\mathrm{Imp}}(\hat x_j;\omega,\pi)+(1-\gamma) \tilde{p} (\hat x_j|y,\pi)\big)} 
    \label{eq:gamma-score}
\end{multline}
\noindent where $\tilde{\mathrm{Imp}}$ and $\tilde{p}$ are the normalized log value of importance weight and backward translation model $p(\hat x|y,\pi)$ as defined in Equation \ref{eq:norm}.

Once the gamma score in Equation \ref{eq:gamma-score} is computed, there are two methods to select $\hat x$ from $A(y)$, which are deterministic and stochastic methods. For deterministic selection, we simply select the candidates with maximum gamma score among $N$ translation candidates; and for sampling, we sample a candidate according to its gamma score distribution. These two methods are called gamma selection and gamma sampling in our experiments.

\section{Experiments}
% In this section we conduct our experiments firstly to explore the existence of quality and importance in NMT back-translation performance. Then we propose Gamma Criterion to balance the weightage between importance and quality in synthetic corpus. 

% We list the above two experiment together in the table \ref{tab:combined} compared with baseline model. 

% We find that, even if we use the combined synthetic corpus, the NMT models trained on synthetic corpus could hardly compare with the baseline whose BLEU score is 32.1. This could be the reason that the bitext is of high quality and importance, and NMT model only need high importance synthetic corpus.

% Therefore, back to the standard back-translation NMT, we conclude that to improve the back-translation NMT models performance, the corpus should firstly consists of sufficient high quality language pairs, which is usually provided by authentic bitext, then provide the NMT model with high importance corpus by synthetic corpus.

% After obtaining the selected language pairs, we combine them with bitext and train the back-translation NMT models. The results for both gamma selection and gamma sampling compared with other back-translation NMT baselines are listed in the table \ref{tab:gamma}.

\subsection{Settings}
We run all the experiments by using fairseq \cite{ott2019fairseq} framework. 
% \lemao{Need to describe datasets: bilingual and monolingual datasets.} 
For dataset settings, since datasets WMT14 EN-DE and DE-EN are widely used~\cite{li2019word,Zhu2020Incorporating,li-etal-2020-evaluating,fan2021mask,le-etal-2021-illinois},  we follow both standard benchmarks and additionally we employ WMT14 RU-EN as the third dataset to validate the effectiveness of the proposed methods. For back translation experiment, we use an equal scale monolingual corpus randomly sampled from Newscrawl 2020 \citep{barrault-etal-2019-findings} comprising 4.5 million monolingual sentences for DE-EN language pair and 2.5 million for RU-EN direction, thus total 9 million sentences for DE-EN pair and 5 million for RU-EN direction are used. We tokenize the parallel corpus using Mose tokenizer \cite{koehn-etal-2007-moses}, and learn a source and target shared Byte-Pair-Encoding (BPE)~\cite{sennrich2016neural} with 32K types. We develop on newstest2013 and report the results on newstest2014.

As for model architecture, we employ all the translation models using architecture $\texttt{transformer\_wmt\_en\_de\_big} $, which is a Big Transformer architecture with 6 blocks in the encoder and decoder, { and is widely used as a standard backbone on various NMT research studies.} 
We use the same hyperparameter settings across all the experiments, i.e., 1024 word representation size, 4096 inner dimensions of feed-forward layers, and dropout is set to 0.3 for all the experiments.
In addition, for monolingual models, we apply $\texttt{transformer\_lm\_gpt}$ architecture \cite{radford2018improving} on source language side of the corpus without any extra corpus.~\footnote{Note that we do not use the pre-trained language models such as GPT-3 or T5 to exclude our gains from large scale monolingual data.} The detailed hyperparameters used for training translation and language models are shown in Appendix. 

\begin{table}
\centering
\resizebox{0.7\linewidth}{!}{%
\begin{tabular}{lcc} \toprule
\multicolumn{1}{c}{\multirow{2}{*}{\textbf{Systems }}} & \multicolumn{2}{c}{\textbf{DE-EN}}       \\ \cmidrule{2-3}
\multicolumn{1}{c}{}                                  & \textbf{w/o bitext} & \textbf{w bitext}  \\ \midrule
Transformer                                              & -                   & 32.1               \\
Beam BT                                                  & 27.6                & 32.7               \\
Sampling BT                                              & 29.2                & 34.1               \\
DM                                                    & \textbf{31.3}       & \textbf{34.2}      \\ \bottomrule
\end{tabular}
}
\begin{tablenotes}
\small
    \item DM means the data manipulation method. 
\end{tablenotes}

\caption{\label{tab:mani_compare} Data manipulation achieves the almost the same BLEU score as sampling BT. }
\end{table}

For baseline models, we train them for 400K updating steps, and train the models with back-translation data for 1.6M updating steps. We save the checkpoints every 100k updating intervals, and only select the checkpoints with highest develop set performance. As for the back-translation data, we study beam decoding and sampling decoding as baselines since they are the common practice for BT research ~\cite{roberts2020decoding, wang2019improving}.
We use baseline models' checkpoints at 400K updating steps to generate default beam5 decoding and sampling decoding synthetic corpus without any penalty. For monolingual models, we only select the checkpoints with the best develop set performance.
When tuning $\gamma$ on dev sets for data manipulation methods we select it from $\{0, 1/4, 1/2, 3/4, 1\}$ and the optimal is $\gamma=1/2$.
For the Gamma Score method, $\gamma$ is tuned among $\{0.1, 0.2, 0.3, 0.4, 0.5\}$ and it is set $\gamma = 0.2$ for all three tasks. 

All the experiments are conducted using 8 Nvidia V100-32GB graphic cards without any gradient accumulation or bitext upsampling, and the results in this paper are measured in case-sensitive detokenized BLEU with SacreBLEU\footnote{We use the fairseq default shell script $\texttt{sacrebleu.sh}$, with WMT14/full testsets to evaluate the model checkpoints. The sacrebleu output format is BLEU + case.mixed + lang.de-en + numrefs.1 + smooth.exp + test.wmt14/full + tok.13a + version.1.4.13.} by \citet{post-2018-call}.

% \begin{table}
% \centering
% \resizebox{\linewidth}{!}{%
% \begin{tabular}{lrrrrr} \toprule
% \multicolumn{1}{c}{\textbf{Combined}} & \multicolumn{1}{c}{\textbf{sampling }} & \multicolumn{1}{c}{\textbf{3:1}} & \multicolumn{1}{c}{\textbf{2:2}} & \multicolumn{1}{c}{\textbf{1:3}} & \multicolumn{1}{c}{\textbf{beam}}  \\ \midrule
% BT BLEU                                       & 7.7                                    & 9.8                              & 13.6                             & 19.2                             & \textbf{27.2}                      \\
% $\log p(\hat{x}|y,\pi)$& -157.62                                & -136.45                          & -104.65                          & -64.13                           & \textbf{-15.65}                    \\
% Mono.                                         & -199.48                                & -184.20                          & -162.72                          & -136.82                          & \textbf{-110.78}                   \\
% Imp.                                          & \textbf{-41.86}                        & -47.75                           & -58.08                           & -72.68                           & -95.13                             \\ \bottomrule
% \end{tabular}
% }
% \caption{\label{tab:ratio_test} The combination of beam and sampling synthetic source sentences on testset. It can be noticed that with the increasing of beam ratio in combined testset, the back-translation BLEU, BT likelihood and monolingual likelihood are increasing, and the importance weights are decreasing.}
% \end{table}

\subsection{Main Results}

\subsubsection{Results on DE-EN}

\paragraph{Data Manipulation}

We conduct two experiments to study the data manipulation for back-translation NMT model performance using aforementioned corpus with and without authentic corpus.

Table \ref{tab:mani_compare} show the data manipulation results compared with baseline. 
Firstly, for synthetic corpus experiment, we find that even if only monolingual corpus is used, the performance of back-translation NMT model can still be significantly improved to 31.3 from 29.2 by sampling or 27.6 by beam, and it is only 0.7 lower than bitext baseline by BLEU score measure. 
% This could be the reason that the beam in combining corpus consists of precise token alignment information, and sampling provides a wide range of learning vocabulary.
Secondly, for the experiments with bitext, the best performance by data manipulation only helps the back-translation NMT model achieves almost the same performance with sampling BT. This means data manipulation methods cannot achieve a higher BLEU score than sampling or beam.

% Finally, by comparing two experiments, we witness a different best-performance point of beam-sampling ratio. For synthetic corpus only, the best performance point is achieved when sampling-beam ratio is ${2:2}$, while it is at ${3:1}$ point in the case without bitext. 
% Such a phenomenon indicates corpus with bitext provides different importance and quality information from the corpus that includes bitext, and the model prefers more corpus of high importance under the bitext settings. However, if only high quality corpus provided (using beam corpus only to train the model), we witness a severe drop of NMT performance. That means the importance weight of the corpus are important.

% We observe that the contribution of data manipulation for settings without bitext is of more significance than the settings with bitext. Comparing with pure sampling and pure beam synthetic corpus, this vanilla manipulating on corpus combination ratio achieves 34.3 BLEU, which is only 0.2 BLEU score higher than pure sampling. The results indicate that, the improvement by corpus manipulation is incremental, and such a \textbf{vanilla corpus manipulation is not sufficient for further improve the back-translation NMT performance.}

\begin{table}
\centering
\resizebox{0.7\linewidth}{!}{%
\begin{tabular}{lc} \toprule
\multicolumn{1}{c}{\textbf{Systems}} & \textbf{SacreBLEU}  \\ \hline
Transformer                          & 32.1                \\
Beam BT                              & 32.7                \\
Sampling BT                          & 34.1                \\
DM +bitext                           & 34.2                \\
Gamma sampling BT                    & \textbf{35.0}*      \\
Gamma selection BT                   & 34.7*               \\ \bottomrule
\end{tabular}
}
\caption{\label{tab:de_en_gamma} BLEU score on WMT14 DE-EN testset. Gamma criterion based method outperform beam search based and sampling based back-translation NMT models. The result marked with * denotes that it is significantly better than sampling BT with $p < 0.0010$.}
\end{table} 

\paragraph{Gamma Score}
% \lemao{key points. 1. the performance of data manipulation with or w/o authentic data and explain why it performs good w/o authentic whereas it delivers modest improvements w authentic data; 2. two implementations of gamma score methods performs better than data manipulation}
In this paragraph, we conduct the experiments based on gamma score method. 
% As aforementioned, there are the deterministic methods and stochastic methods for sampling candidate. 
We conduct both of the methods in this experiment: we select the candidate with highest gamma score for the deterministic method whereas sample the candidate by gamma score distribution for the stochastic method. 
\begin{table}
\centering
%\resizebox{1\linewidth}{!}
{%
\begin{tabular}{l%!{\vrule width \lightrulewidth} 
ccc} \toprule
\multicolumn{1}{c}{\textbf{System}}& \textbf{EN-DE} & \textbf{RU-EN}  \\ \hline
Transformer                                                & 27.4           & 34.1            \\
Beam BT                                                  & 29.7           & 35.9            \\
Sampling BT                                            & 30.0           & 35.6            \\
Gamma selection BT                       & \textbf{31.0}*  & 36.1*            \\ 
Gamma sampling BT                   & 30.9*           & \textbf{36.3}*   \\
\bottomrule
\end{tabular}}
\caption{\label{tab:gamma} 
SacreBLEU score on WMT14 EN-DE and RU-EN testsets. Gamma criterion based methods outperform beam search based and sampling based back-translation NMT models. The result marked with * denotes that it is significantly better than both sampling and beam based BT with $p<0.001$.}
\end{table}

\begin{table*}
    \centering
    \resizebox{\linewidth}{!}{%
    \begin{tabular}{ccc}
     \includegraphics[width=0.3\textwidth]{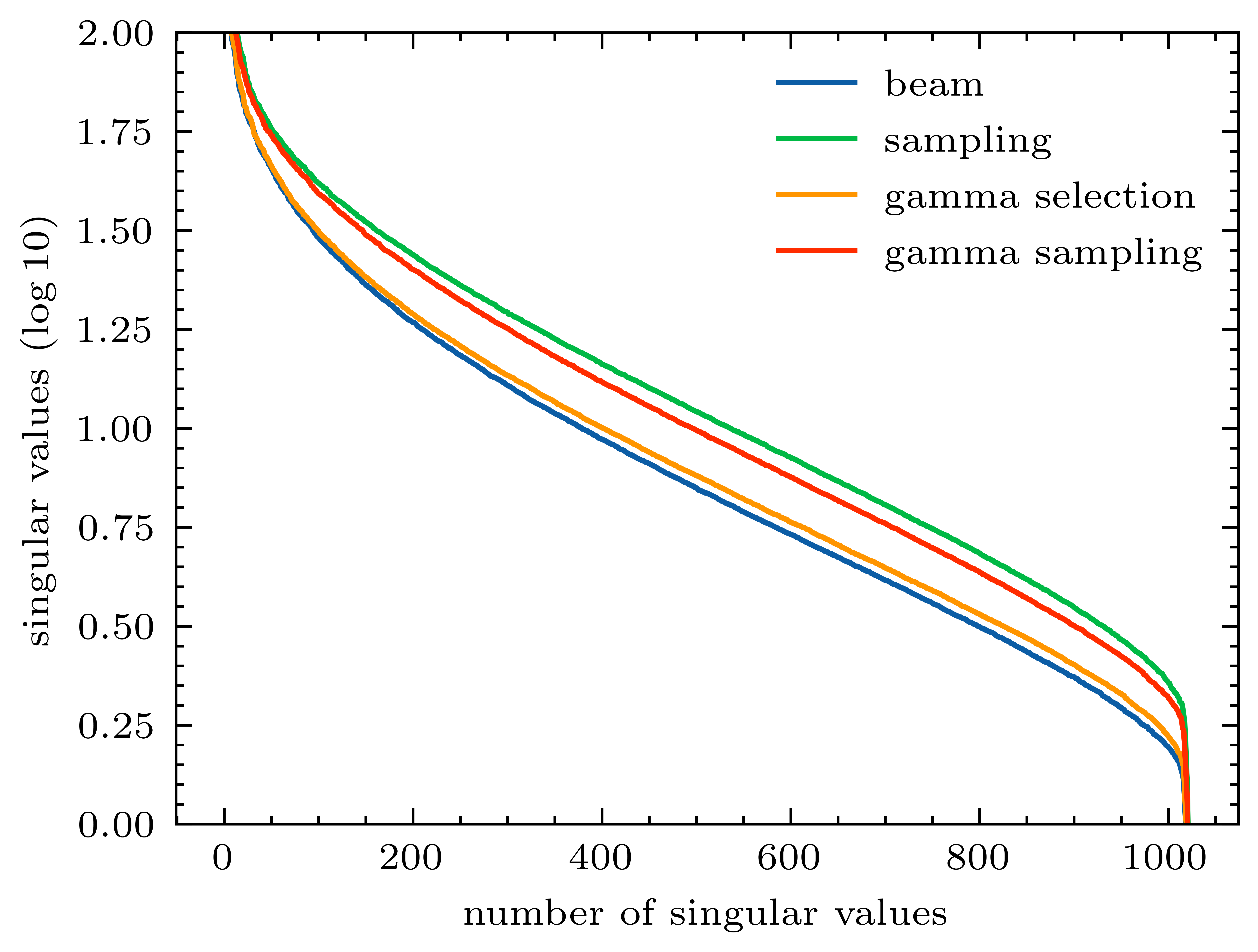} &\includegraphics[width=0.3\linewidth]{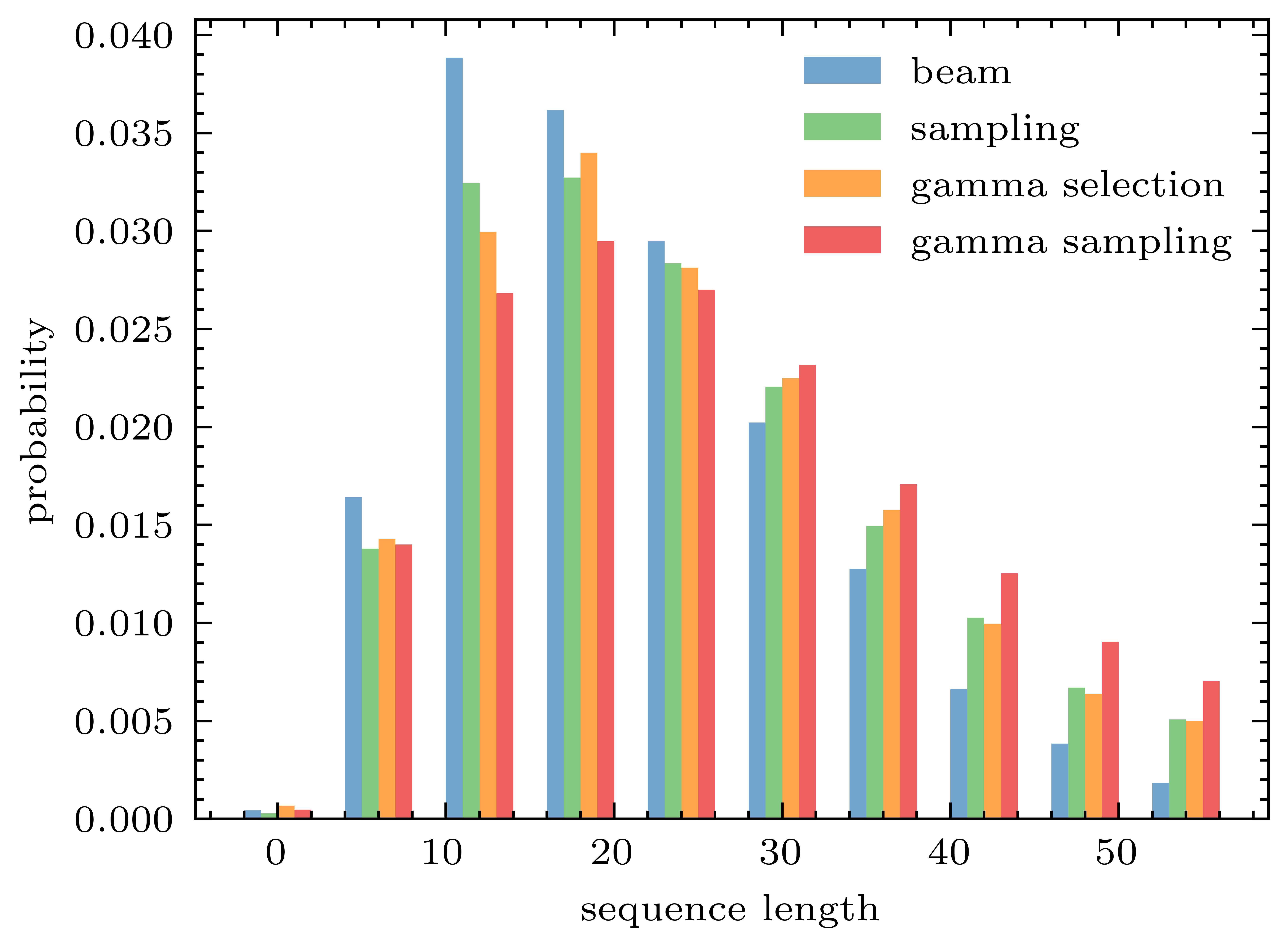}  & \includegraphics[width=0.3\linewidth]{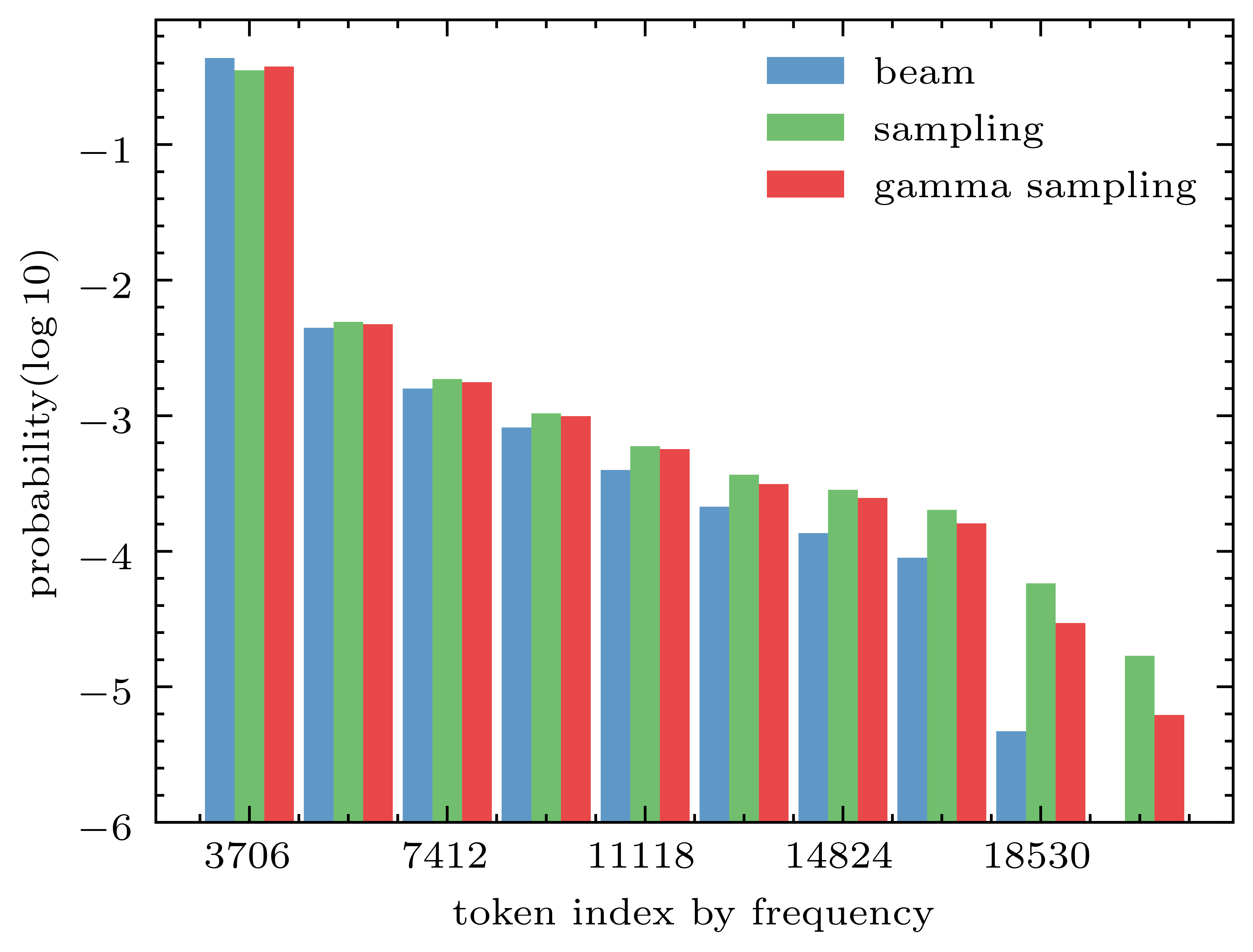}  \\
     \small (a) Spectrum & \small (b) Sequence Length  & \small (c) Token frequency 
    \end{tabular}
    }
    \captionof{figure}{\label{graph:analysis} Synthetic corpus analysis on singular value spectrum(a), sequence length histogram(b) and token frequency histogram(c).}
\end{table*}

Once again, we use synthetic gamma corpus combined with bitext to train the back-translation NMT models on each corpus, the results are listed in \ref{tab:de_en_gamma}. From the table, we can see that our proposed gamma sampling significantly outperforms the sampling based and beam search based back-translation baselines by 0.9 and 2.3 BLEU scores in terms of SacreBLEU. And our two proposed gamma score based methods outperform the data manipulation method as well.

% \paragraph{Effects of $\gamma$ in Gamma Score}
% \lemao{You need to present results on dev sets}
% As a weightage factor in our proposed gamma score methods, the optimum value of $\gamma$ could be tuned. However, 
% Since the motivation is to prove the effectiveness of our proposed methods, here we just employ $\gamma=0.2$ across all the experiments. The details of the gamma search could be found in the appendix. 

In the rest of the experiments, we report results for both gamma selection and gamma sampling as the proposed methods and their hyperparameter $\gamma$ for other tasks is fixed to $0.2$. 
\subsection{Results on other Datasets}
% \lemao{key point: present the results of gamma score on all other datasets}
We conduct the experiments on WMT14 EN-DE and RU-EN for both gamma selection and gamma sampling as well, and table \ref{tab:gamma} shows that our proposed gamma based methods significantly outperform beam and sampling based back-translation methods on both en-de and ru-en translation for almost 1 and 0.4 BLEU score respectively. Recently, \citet{edunov2020evaluation} point out that BLEU might overlook the contributions from back translation since it poorly correlates with human evaluation on the data generated in back translation scenario. Follow their suggestions, to better reflect the scenario of back translation, we also evaluate our experiment using COMET metric suggested by \citet{rei2020comet}. The results are shown in table \ref{tab:comet} and we can see that the proposed methods perform well in terms of COMET.

\begin{table}
\centering
\resizebox{\linewidth}{!}{%
\begin{tabular}{lccc} \toprule
\multicolumn{1}{c}{\textbf{System}} & \textbf{DE-EN} & \textbf{EN-DE} & \textbf{RU-EN} \\ \midrule
Transformer & 51.66 & 53.35 & 54.55 \\
Beam BT & 49.35 & 54.61 & 55.12 \\
Sampling BT & 52.71 & 56.01 & 54.34 \\
Gamma Selection BT & 53.83 & \textbf{58.22} & \textbf{57.03} \\
Gamma Sampling BT & \textbf{53.97} & 58.18 & 56.69 \\ \bottomrule
\end{tabular}
}
\caption{\label{tab:comet}
COMET metric evaluation results on WMT14 DE-EN, EN-DE and RU-EN datasets. The testset results are in accordance with BLEU metric.
}
\end{table}

\paragraph{Discussion on Efficiency}
Since our method requires to run sampling with size of 50 to generate synthetic data, its efficiency is about 10x slower than that of beam BT with size of 5 and 50x slower than that of sampling BT with size 1. Luckily, because the bottleneck of BT is not the synthetic data generation but the parameter optimization on both synthetic and authentic data, our overall overhead is less than 0.5x slower than sampling BT. In addition, since decoding is very easy to be parallelized on GPU or CPU machines, the cost of decoding is not a serious issue for our method, which makes it possible to run our method on a large scale dataset.

\subsection{Analysis on Synthetic Corpus}
In this subsection, we analyze the synthetic corpus of proposed gamma score methods on both sentence level and token level.

\paragraph{Sentence Level}
We evaluate the back-translation synthetic source sentences by their sentence representations. 
% Firstly, we generate the synthetic source by beam, sampling, gamma selection, and gamma sampling four strategies. 
We use the baseline model to generate the hidden representations at the
% . Here, we use the last hidden state 
end-of-speech token as the sentence representation. Here, we compute the singular value spectrum of the representations for different back-translation corpora. \footnote{Singular value spectrum analysis is a widely used method to measure the representation distribution. \citet{gao2018representation} firstly introduces this method to measure the isotropy of representation, and \citet{wang2019improving} directly employ spectrum control for better NMT performance. The idea is, representations of high linguistic variety usually are more isotropic, thus having a relatively uniform singular value distribution. We employ this method here to measure the variety of sentence-level information. }

The spectrum is shown in figure \ref{graph:analysis}(a). From the spectrum, sampling has a more uniform distribution whereas beam has the worst variety. Our proposed methods have moderate variety between sampling and beam, and gamma sampling consists of higher linguistic information richness compared with gamma selection. 

Figure \ref{graph:analysis}(b) shows the sequence length of the synthetic corpora of different generation methods. Beam generates the shortest synthetic sentences and gamma sampling generates the longest synthetic sentences on average. Between them, sampling and gamma selection generate almost the same sequence length, which means gamma selection candidates provide more learning signals than random sampling under the same length.

% computing the ppl 5-gram model, on testset of back translated beam, sampling, gamma selection, gamma sampling, and also the forward sys prediction

\paragraph{Token Level}

% We also count the token level analysis between our proposed gamma sampling with beam and sampling. 
% We separate the vocabulary into 10 equal size intervals by counting token frequency training set, and compute the probability for each token falls into the intervals in synthetic corpus. 
Figure \ref{graph:analysis}(c) is the token frequency histogram, which shows beam has higher probability to decode high frequency tokens while sampling prefers more low frequency tokens. 
% sampling and gamma sampling has more low frequency tokens compared with beam.

We also measure the vocabulary size, finding that the proposed gamma sampling shares the same vocabulary size as sampling method. This could be the reason that gamma sampling is based on random sampling for candidates generation. 
% and the same vocabulary size as random sampling ensures the variety of candidates generated, thus to further improve NMT model test performance. 

% \begin{figure}[htbp]
% \centering
% \includegraphics[width=0.47\textwidth]{masked_token.png}
% \caption{Accumulating number of masked tokens by token frequency. We mask 20\% tokens in each candidates for random masking, masking tokens with highest quality BT likelihood and masking tokens with highest importance likelihood. We could infer which part of frequency tokens are mostly masked by function slop. It can be concluded that the tokens with high quality are mostly high frequency tokens and the high importance tokens are of low frequency. Thus, to trade-off the candidates quality and importance in token level is likely to squeeze the candidates to natural token distribution as shown in random masking line.}
% \label{graph:token_masking}
% \end{figure}

\section{Related Work}
This section describes prior arts in back-translation for NMT,  data augmentation, and semi-supervised machine translation.

\paragraph{Back-translation NMT}
\citet{bojar2011improving} firstly proposed back-translation, then \citet{bertoldi2009domain, lambert2011investigations} apply back translation to solve the domain adaptation problems in phrase-based NMT systems.  \citet{sennrich-etal-2016-improving} further extend the back translation for training NMT models integrally.
%and find that auto back translation as target side 
%corpus could boost the fluency for phrase-based statistical machine translation. 

For understanding the back-translation synthetic corpus, \citet{currey-etal-2017-copied} use a copy of target as a pseudo source, and find that NMT model performance can still be improved under the low resource settings. \citet{caswell2019tagged} propose tagged back-translation to indicate to the model that the given source is synthetic. To further find an optimum back-translation corpus decoding method, \citet{imamura2018enhancement} firstly use sampling based synthetic corpus and find such a stochastic decoding method outperforms beam search on boosting NMT model performance, and \citet{edunov2018understanding} broaden the investigation of a number of back-translation generation methods for synthetic source sentences. Their contribution shows that sampling or noisy synthetic data gives a much stronger training signal. \citet{gracca2019generalizing} reformulate back-translation in the context of optimization and clarifying to improve sampling based decoding method search space, thus proposing N best list sampling. Recently, \citet{nguyen2020data} diversify the training data by multiple forward and backward models translations and combine them with the original datasets.

% For back-translation training objective studies, \citet{fadaee2018back} shows that training a NMT systems with high prediction loss words benefit most from the addition of synthetic data.  At the same time, \citet{hoang2018iterative} provides insights into back-translation sampling and weighting strategy and generalize it to iterative back-translation models. And recently, \citet{wang2019improving} proposes methods to measure the confidence of NMT model predictions based on uncertainty, which further cope with noise in synthetic bilingual corpora.  \citet{jiao2021alternated} introduce authentic data as guidance to prevent the training of NMT models from being disturbed by noisy synthetic data.

\paragraph{Data Augmentation for NMT}
NMT researchers are the pioneers of data augmentation studies since back-translation is a natural type of data augmentation method. \cite{sennrich-etal-2016-improving, NIPS2016_2f885d0f, zhang2016exploiting, bi-etal-2021-data}.
To balance the token frequency in NMT corpus,
\citet{fadaee2017data} create new sentences contain low-frequency words. However, as observed by \citet{wang2018switchout}, the improvement across different translation tasks is not consistent, and they invent SwitchOut data augmentation policy. \citet{recht2018cifar, recht2019imagenet, werpachowski2019detecting} also observe such an inconsistency of variance between training corpus and testing set as well as in the generation tasks.
Recently, \citet{li2019understanding} try to understand data augmentation from input sensitivity and prediction margin, thus obtaining relatively low variance in generation. 
% \lemao{This paper refers many papers about this direction, see Understanding Data Augmentation in Neural Machine Translation:Two Perspectives towards Generalization}

\paragraph{Semi-supervised Machine Translation}
However, as high quality bitext is always limited and costly to collect, \citet{gulcehre2015using} study methods for effectively leveraging monolingual data in NMT systems. \citet{NIPS2016_5b69b9cb} develop a dual-learning mechanism, under such a learning objective, a NMT system is able to automatically learn from unlabeled data, thus improving NMT performance iteratively. Based on iterative learning, \citet{lample2018phrase} investigates how to learn NMT systems when only large monolingual corpora can be used in each language. 

For supervision of models, \citet{gulcehre2017integrating} employ the target language model hidden states into NMT decoder to further improve performance.
\citet{edunov2020evaluation} show that back-translation improves translation quality of both naturally occurring text and translationese according to professional human translators. 
For supervision of learning corpus, \citet{wu2019exploiting} study both the source-side and target-side monolingual data for NMT.

% \lemao{Exploiting monolingual data at scale for neural machine translation; Exploiting monolingual data at scale for neural machine translation.}
\section{Conclusion}
In this work, we answer a fundamental question about synthetic data for back translation. We theoretically and empirically show two key factors namely quality and importance weight of synthetic data play an important role in back translation, and then we propose a new method to generate synthetic data which better balances both factors so as to boost the back-translation performance. For future work, we think it would be of significance to apply our synthetic data generation method to other BT methods or even to more broad NLP tasks such as paraphrasing and style transfer. 

\section*{Acknowledgements}
We would like to thank anonymous (meta) reviewers for suggestions on this paper. 
L. Liu is the corresponding author. 

\bibliography{custom}
\bibliographystyle{acl_natbib}

\appendix
\section{Model Details}
The models are optimized using Adam optimizer \cite{kingma2015adam}, with $\beta_1=0.9, \beta_2=0.98$. We use the same learning rate schedular as \cite{vaswani2017attention} with maximum learning rate $7\times10^{-4}$, and 4000 warmup updates. 
We use the fairseq 10.2 as the framework and the training command as well as the model hyperparameters are listed below,
\begin{verbatim}
fairseq-train \
--arch transformer_wmt_en_de_big 
--share-all-embeddings 
--dropout 0.3
--weight-decay 0.0
--criterion 
    label_smoothed_cross_entropy
--label-smoothing 0.1
--optimizer adam 
--adam-betas '(0.9, 0.98)' 
--clip-norm 0.0
--lr-scheduler inverse_sqrt 
--warmup-updates 4000
--max-tokens 4096 
--max-update 1600000
--validate-interval-updates 10000
--save-interval-updates 100000
--lr 7e-4 
--upsample-primary 1
\end{verbatim}

{And the GPT model we employ is only trained on source side of bitext corpus, without extra datasets. The training command line and core settings are listed below.}
\begin{verbatim}
fairseq-train \
--task language_modeling 
--arch transformer_lm_gpt 
--share-decoder-input-output-embed 
--dropout 0.1
--optimizer adam 
--adam-betas '(0.9, 0.98)' 
--weight-decay 0.01 
--clip-norm 0.0
--lr 7e-5 
--lr-scheduler inverse_sqrt 
--warmup-updates 8000
--tokens-per-sample 512 
--sample-break-mode none
--max-tokens 4096 
--update-freq 1
--max-update 1000000
--keep-last-epochs 5
--validate-interval-updates 10000
--save-interval-updates 10000
\end{verbatim}

% \section{Gamma Search}
% In this section, we run the gamma hyperparameter search. The performance on develop set of gamma selection back-translation models on each gamma settings are in the figure \ref{graph:gamma_search}.
% \begin{figure}[h]
% \centering
% \includegraphics[width=0.47\textwidth]{gamma_search.png}
% \caption{\label{graph:gamma_search} $\gamma$ search on develop set.}
% \end{figure}
% It can be noticed that, simple $gamma$ weightage on importance weights can continuously improve $\gamma$ score based back-translation NMT model performance. However, it could be difficult to search the $\gamma$ optimum point since it is very sensitive to the experiment settings. Therefore, we conduct all the experimnet based on $\gamma=0.2$, and such a $\gamma$ setting still works well on other datasets.
% This is an appendix.

% \section{Jessen Inequality}
% \xu{Do we need this? }
% In this section we justify the Jessen inequality we use. We use Talyer series of log algorithm to estimate the Jessen inequality difference.
% \begin{align}
%     \log(x)=x-1+o((x-1)^2)
% \end{align}
% Difference between before and after scaling of Jensen's inequality is,
% \begin{align}
%     \Delta&=\log\sum_x p(x)p(y|x)-\sum_x p(x)\log p(y|x)\nonumber\\
%     &=\sum_x p(x)p(y|x) -1 -\sum_x p(x)(p(y|x)-1)\nonumber\\
%     &+o(\sum_x p(x)p(y|x)-1)^2-\sum_x p(x)o(p(y|x)-1)^2 \nonumber\\
%     &=o(p(y)-1)^2-\mathbb{E}_x\{o(p(y|x)-1)^2\}
% \end{align}
% Based on our analysis, the inequality difference is a second order infinitesimal.

\end{document}